\documentclass[11pt]{article}

\usepackage{pgfplots}
\pgfplotsset{compat=1.17}
\usepackage[final]{acl}
\usepackage{pgfplots}
\usepackage{times}
\usepackage{latexsym}
\usetikzlibrary{patterns}
\usepackage{color}
\usepackage{xcolor}
\usepackage{subcaption}

\usepackage[dvipsnames]{xcolor}
\usepackage[T1]{fontenc}

\usepackage[utf8]{inputenc}

\usepackage{microtype}

\usepackage{inconsolata}

\usepackage{graphicx}
\usepackage{booktabs} 
\usepackage{amsmath}  
\usepackage{pgfplots}
\pgfplotsset{compat=1.17}
\usepgfplotslibrary{groupplots} 
\usepgfplotslibrary{colormaps}  
\title{Do Mixed-Vendor Multi-Agent LLMs Improve Clinical Diagnosis?}

\author{
  Grace Chang Yuan$^{1}$ \quad
  Xiaoman Zhang$^{2}$ \quad
  Sung Eun Kim$^{2}$ \quad
  Pranav Rajpurkar$^{2}$\thanks{Corresponding author} \\
  $^{1}$ Massachusetts Institute of Technology, Boston, MA \\
  $^{2}$ Department of Biomedical Informatics, Harvard Medical School, Boston, MA
}

\begin{document}
\maketitle

\begin{abstract}
Multi-agent large language model (LLM) systems have emerged as a promising approach for clinical diagnosis, leveraging collaboration among agents to refine medical reasoning. However, most existing frameworks rely on single-vendor teams (e.g., multiple agents from the same model family), which risk correlated failure modes that reinforce shared biases rather than correcting them. We investigate the impact of vendor diversity by comparing Single-LLM, Single-Vendor, and Mixed-Vendor Multi-Agent Conversation (MAC) frameworks. Using three doctor agents instantiated with o4-mini, Gemini-2.5-Pro, and Claude-4.5-Sonnet, we evaluate performance on RareBench and DiagnosisArena. Mixed-vendor configurations consistently outperform single-vendor counterparts, achieving state-of-the-art recall and accuracy. Overlap analysis reveals the underlying mechanism: mixed-vendor teams pool complementary inductive biases, surfacing correct diagnoses that individual models or homogeneous teams collectively miss. These results highlight vendor diversity as a key design principle for robust clinical diagnostic systems. Our code is publicly available.\footnote{\url{https://github.com/rajpurkarlab/mixed-vendor-mac}}
\end{abstract}

\section{Introduction}
Large language models are increasingly used for clinical diagnosis \citep{kim2024mdagents, lee2025soap, nori2025mai, zhou2025mam}. Across a range of benchmarks, frontier models from OpenAI, Google, and Anthropic can approach or even exceed average physician performance \citep{nori2025mai, saab2024capabilities}. Yet no single model dominates: performance varies substantially across tasks and clinical domains, with different vendors exhibiting distinct strengths and weaknesses \citep{cao2024liver, sonoda2024radiology}.

This variability has motivated the development of multi-agent frameworks, where several LLM ``doctors'' collaborate under a coordinating supervisor \citep{kim2024mdagents, lee2025soap, zhou2025mam}. Such systems more closely resemble real clinical workflows, multidisciplinary case conferences where diverse perspectives converge on a diagnosis \citep{chen2025mac, zhou2025mam}. Early results are promising, with such architectures frequently outperforming single-model baselines \citep{chen2025mac, nori2025mai, zhou2025mam}.

A small body of work has begun exploring mixed-vendor systems in general domains \citep{ye2025xmas}. However, existing mixed-vendor approaches in medicine mostly rely on independent model runs, combining models from different providers via voting or simple aggregation \citep{barabucci2024combining, sorich2024triage}. Research integrating cross-vendor models within structured multi-agent conversations to directly compare them against single-vendor baselines remains limited.

We address this gap by hypothesizing that vendor diversity improves diagnosis by introducing complementary inductive biases and breaking the ``echo chamber'' of homogeneous teams. We test this hypothesis using the Multi-Agent Conversation (MAC) framework \citep{chen2025mac}, comparing single-model, single-vendor MAC, and mixed-vendor MAC configurations across RareBench (rare diseases) \citep{chen2024rarebench} and DiagnosisArena (complex case reports) \citep{ zhu2025diagnosis}.

Our results demonstrate that \textbf{mixed-vendor configurations deliver superior overall performance compared to their single-vendor counterparts}, achieving state-of-the-art recall and accuracy across both benchmarks. 
Analysis of solution space overlap reveals the mechanism driving this gain: while single-vendor teams remain highly correlated with their base model's priors, \textbf{mixed-vendor teams successfully pool complementary inductive biases to surface diagnoses that individual models miss}. These findings suggest that leveraging the complementary strengths of diverse LLMs is crucial for building robust clinical diagnostic systems.

\section{Related Work}

\paragraph{Clinical LLMs and Vendor Differences}
Single LLM systems have demonstrated impressive capabilities in medical reasoning, often matching or even exceeding average physician performance \citep{eriksen2024, singhal2025, sonoda2024radiology}. However, these models exhibit divergent behaviors driven by differences in pre-training data and alignment strategies \citep{chen2025sycophancy, li2025alignment}. Rather than having a universally superior model, existing research reveals that models from different vendors possess complementary inductive biases \citep{yang2024onellm}. Consequently, there is potential in employing vendor diversity to bridge these domain-specific gaps, combining distinct reasoning behaviors to achieve more robust clinical performance.

\paragraph{Multi-agent LLM Frameworks in Clinical Diagnosis}
Beyond single-LLM systems, several recent works organize multiple LLM “doctors” into coordinated teams that more closely resemble clinical workflows. Some architectures use a supervisor–specialist pattern, where a director LLM coordinates a set of role-specialized agents and integrates their outputs into a final decision \citep{kim2024mdagents, zhou2025mam}. Others emphasize conversational debate, with multiple specialists independently proposing hypotheses, critiquing one another, and iterating under the oversight of a manager agent \citep{chen2025mac, lee2025soap, nori2025mai}. Another type of multi-agent framework uses panel-style ensembles, where the physician agents reason in parallel with limited interaction and their responses are aggregated into a single collective output \citep{barabucci2024combining}.

\paragraph{Mixed-vendor Multi-agent Systems}
In contrast to the extensive work on single-vendor multi-agent frameworks, mixed-vendor systems have received limited attention. Existing work mostly relies on ``panel-style" aggregation rather than conversational collaboration. 
For example, recent studies demonstrate that combining outputs from cross-vendor models via majority voting or collective intelligence algorithms can outperform individual baselines \citep{barabucci2024combining, sorich2024triage}. However, these frameworks treat agents as independent voters, missing the chance for agents to cross-examine and refine opposing views in real-time. More importantly, the clinical literature lacks a systematic comparison between single-vendor and mixed-vendor configurations within a shared conversational architecture. This raises the question of whether mixed-vendor teams can introduce reasoning diversity that improves diagnostic performance over single-vendor baselines. 

\section{Methods}

\subsection{MAC Framework}
\begin{figure*}[h!]
    \centering
    \includegraphics[width=\linewidth]{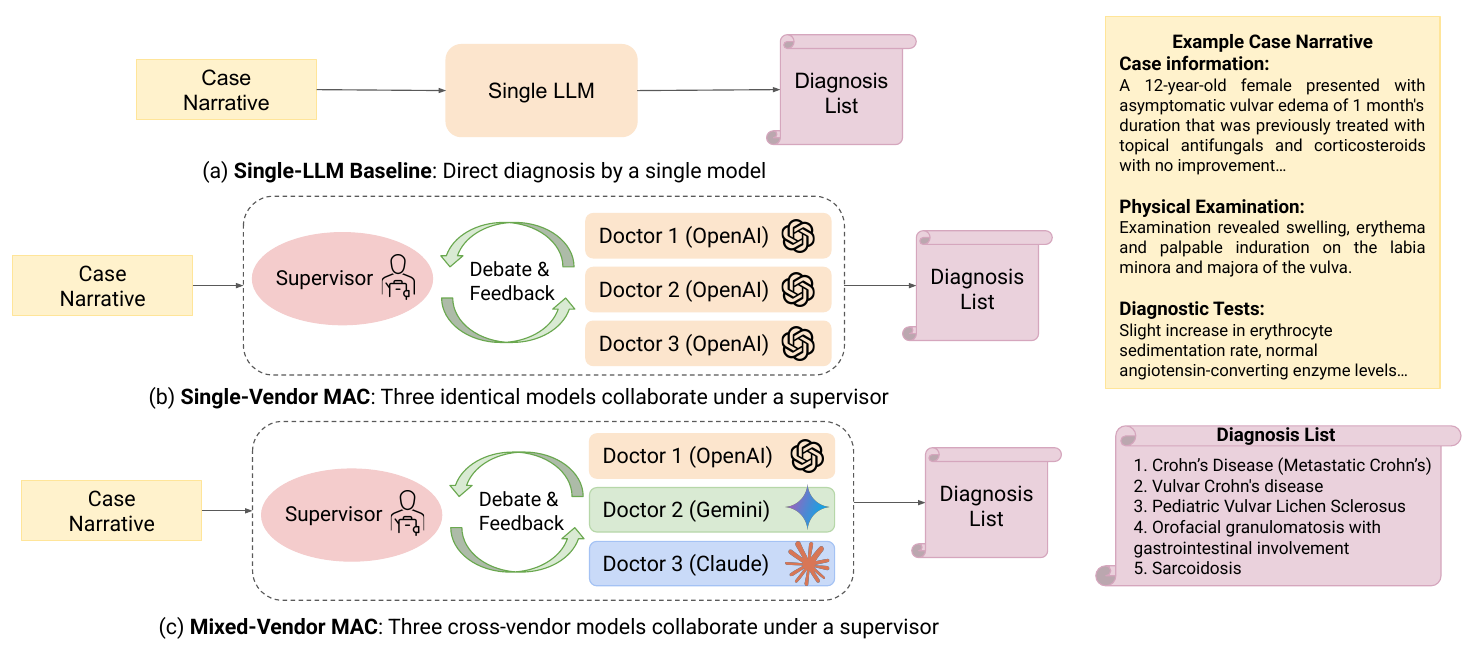}
    \caption{Overview of the 3 different model configurations (Single-LLM, Single-vendor MAC, and Mixed-vendor MAC), and an illustrative example of a case narrative with its corresponding diagnosis list.}
    \label{fig:placeholder}
\end{figure*}
We build directly on the Multi-Agent Conversation (MAC) framework \citep{chen2025mac}, which structures clinical decision-making as a dialogue between three ``doctor" agents and one supervising agent. The process follows a fixed protocol: each doctor LLM proposes and revises ranked diagnosis lists in order, while the supervisor LLM monitors the discussion, provides feedback, and ultimately decides on the final output. Unlike \citet{chen2025mac}, we do not use role-specialized prompts. Instead, all doctor agents receive the same generic clinical instructions adapted to each benchmark. This simple, symmetric setup allows us to vary the underlying LLMs while keeping the interaction pattern unchanged, ensuring that any observed performance differences come strictly from vendor diversity rather than artifacts of role assignment.

We compare three specific model configurations, as shown in (Figure~\ref{fig:placeholder}):
\begin{itemize}
    \item \textbf{Single-LLM}, where a single model (o4-mini, Gemini-2.5-Pro, Claude-4.5-Sonnet) directly produces a diagnosis list for each case.  
    \item \textbf{Single-vendor MAC}, where all three doctor agents are instantiated with the same model under a fixed supervisor.
    \item \textbf{Mixed-vendor MAC}, where the three doctor agents are instantiated with different vendor models under a fixed supervisor.
\end{itemize}

\paragraph{Model Selection} For all experiments, we employ the current frontier models from each vendor: o4-mini (2025-04-16) \citep{openai2025o4}, Gemini-2.5-Pro \citep{google2025gemini}, and Claude-4.5-Sonnet \citep{anthropic2025claude}. 
Preliminary experiments showed comparable diagnostic performance across models, so we treat them as matched frontier baselines and focus on the effects of multi-agent composition. Unless otherwise noted, o4-mini serves as a fixed supervisor to standardize aggregation.

\subsection{Conversation Protocol and Stopping Rule}
For each case, the MAC framework receives a phenotype summary or case narrative. The supervisor initiates the discussion, and the three doctor agents contribute in a fixed round-robin order. At each turn, doctor agents review the full conversation history and the supervisor’s latest feedback before submitting a ranked Top-10 diagnosis list with brief justification.
The conversation terminates when the supervisor determines the differential has stabilized, or upon reaching a hard limit of 13 turns (following the original MAC protocol). We extract the final consensus Top-10 list from the supervisor’s concluding message using a rule-based parser and evaluate it against the ground truth.

\section{Experimental Setup}

\subsection{Datasets}

We evaluate on two distinct benchmarks to test generalizability across rare and common conditions. 

\paragraph{RareBench} Designed to evaluate the capabilities of LLMs on rare diseases, RareBench \citep{chen2024rarebench} consists of multiple subsets with varing difficulty. We utilize \textbf{MME} (40 cases across 17 diseases), \textbf{HMS} (88 cases covering 39 diseases), and \textbf{LIRICAL} (370 cases spanning 252 diseases). 

\paragraph{DiagnosisArena} To test performance on broader clinical reasoning, we include DiagnosisArena, a benchmark of 1,113 complex case reports from 10 top-tier medical journals. 
Even strong reasoning models such as o3-mini, o1, and DeepSeek-R1 achieve at most about 46\% accuracy on this benchmark \citep{zhu2025diagnosis}, highlighting its difficulty. 
To minimize the risk of data contamination and ensure computational feasibility, we restrict our evaluation to the subset of 165 cases published in 2024. This temporal restriction reduces overlap with model pre-training data and provides a cleaner test of generalization. 

\subsection{Tasks and Metrics}

\paragraph{RareBench} For this benchmark, the task is to output a ranked Top-10 diagnosis list for each case. We evaluate diagnostic performance using Recall@1, Recall@3, Recall@5, and Recall@10.

\paragraph{DiagnosisArena} In this setting, the task is to generate a ranked Top-5 diagnosis list for each case. We assess performance using Top-1 and Top-5 accuracy, consistent with the standard evaluation protocol for this benchmark. 

 \paragraph{Evaluation Protocols} Our primary evaluation uses o4-mini as an LLM judge, scoring free-text differential diagnoses against reference labels. To reduce potential self-preference bias, we also implement a vendor-neutral, retrieval-based judge: BioLORD (FRemyCompany/BioLORD-2023) \citep{remy2024biolord}. In this secondary evaluation, we embed the predicted diagnoses and match them to the closest standardized code before comparing with the gold code.

\begin{table*}[t!]
    \centering
    \small
    \setlength{\tabcolsep}{4pt} 
    
    \begin{minipage}[t]{0.48\textwidth}
        \centering
        \caption{Performance on \textbf{Combined RareBench}. Note: OpenAI refers to o4-mini, Gemini refers to Gemini-2.5-Pro, and Claude refers to Claude-4.5-Sonnet.}
        \label{tab:combined}
        \resizebox{\linewidth}{!}{
            \begin{tabular}{lrrrr}
            \toprule
            \textbf{Setup} & \textbf{R@1} & \textbf{R@3} & \textbf{R@5} & \textbf{R@10} \\
            \midrule
            Single-LLM (OpenAI) & 31.81 & 40.52 & 46.12 & 53.07 \\
            Single-LLM (Gemini) & 37.58 & 47.79 & 50.84 & 56.41 \\
            Single-LLM (Claude) & 29.35 & 39.54 & 45.41 & 55.42 \\
            Single-Vendor MAC (OpenAI) & 32.73 & 41.22 & 45.54 & 52.33 \\
            Single-Vendor MAC (Gemini) & 36.63 & 43.43 & 46.47 & 51.93 \\
            Single-Vendor MAC (Claude) & 33.32 & 48.06 & 50.53 & 57.06 \\
            \textbf{Mixed-Vendor MAC} & \textbf{39.31} & \textbf{49.82} & \textbf{55.05} & \textbf{61.35} \\
            \bottomrule
            \end{tabular}
        }
    \end{minipage}
    \hfill
    \begin{minipage}[t]{0.48\textwidth}
        \centering
        \caption{Performance on \textbf{MME} subset (n=40). Model versions consistent with Table \ref{tab:combined}.}
        \label{tab:mme}
        \resizebox{\linewidth}{!}{
            \begin{tabular}{lrrrr}
            \toprule
            \textbf{Setup} & \textbf{R@1} & \textbf{R@3} & \textbf{R@5} & \textbf{R@10} \\
            \midrule
            Single-LLM (OpenAI) & 10.00 & 22.50 & 25.00 & 27.50 \\
            Single-LLM (Gemini) & 30.00 & 37.50 & 42.50 & 47.50 \\
            Single-LLM (Claude) & 2.50 & 22.50 & 30.00 & 42.50 \\
            Single-Vendor MAC (OpenAI) & 15.00 & 25.00 & 30.00 & 35.00 \\
            Single-Vendor MAC (Gemini) & 35.00 & 45.00 & 47.50 & 47.50 \\
            Single-Vendor MAC (Claude) & 35.00 & 45.00 & 50.00 & 57.50 \\
            \textbf{Mixed-Vendor MAC} & \textbf{40.00} & \textbf{52.50} & \textbf{60.00} & \textbf{65.00} \\
            \bottomrule
            \end{tabular}
        }
    \end{minipage}

    \vspace{0.6cm} 

    \begin{minipage}[t]{0.48\textwidth}
        \centering
        \caption{Performance on \textbf{HMS} subset (n=88). Model versions consistent with Table \ref{tab:combined}.}
        \label{tab:hms}
        \resizebox{\linewidth}{!}{
            \begin{tabular}{lrrrr}
            \toprule
            \textbf{Setup} & \textbf{R@1} & \textbf{R@3} & \textbf{R@5} & \textbf{R@10} \\
            \midrule
            Single-LLM (OpenAI) & 43.18 & 57.95 & \textbf{68.18} & 76.14 \\
            Single-LLM (Gemini) & 47.73 & 59.09 & 64.77 & 70.45 \\
            Single-LLM (Claude) & 39.77 & 56.82 & 65.91 & 75.00 \\
            Single-Vendor MAC (OpenAI) & 47.73 & 56.82 & 62.50 & \textbf{77.27} \\
            Single-Vendor MAC (Gemini) & 43.18 & 51.14 & 56.82 & 59.09 \\
            Single-Vendor MAC (Claude) & 43.18 & 56.82 & 67.05 & 72.73 \\
            \textbf{Mixed-Vendor MAC} & \textbf{51.14} & \textbf{60.23} & \textbf{68.18} & \textbf{77.27} \\
            \bottomrule
            \end{tabular}
        }
    \end{minipage}
    \hfill
    \begin{minipage}[t]{0.48\textwidth}
        \centering
        \caption{Performance on \textbf{LIRICAL} subset (n=370). Model versions consistent with Table \ref{tab:combined}.}
        \label{tab:lirical}
        \resizebox{\linewidth}{!}{
            \begin{tabular}{lrrrr}
            \toprule
            \textbf{Setup} & \textbf{R@1} & \textbf{R@3} & \textbf{R@5} & \textbf{R@10} \\
            \midrule
            Single-LLM (OpenAI) & 28.61 & 36.51 & 41.14 & 48.50 \\
            Single-LLM (Gemini) & \textbf{36.24} & 45.78 & 48.50 & 53.95 \\
            Single-LLM (Claude) & 29.43 & 40.33 & 43.87 & 52.32 \\
            Single-Vendor MAC (OpenAI) & 30.25 & 38.96 & 41.96 & 47.41 \\
            Single-Vendor MAC (Gemini) & 33.79 & 41.42 & 43.87 & 50.95 \\
            Single-Vendor MAC (Claude) & 31.88 & 43.60 & 46.59 & 52.59 \\
            \textbf{Mixed-Vendor MAC} & 35.69 & \textbf{46.59} & \textbf{51.77} & \textbf{57.22} \\
            \bottomrule
            \end{tabular}
        }
    \end{minipage}
\end{table*}

\section{Results}

\subsection{RareBench}

We present results on RareBench in Tables~\ref{tab:combined}--\ref{tab:lirical}, covering aggregate performance on the MME, LIRICAL, and HMS subsets. Table~\ref{tab:combined} demonstrates that the Mixed-Vendor MAC achieves the highest performance across all retrieval metrics, securing a Recall@1 (39.31\%) and maintaining a clear lead through Recall@10 (61.35\%). Crucially, it consistently outperforms every Single-Vendor MAC baseline. While individual models fluctuate in performance across subsets (e.g., OpenAI struggles on MME but excels on HMS), the mixed-vendor team effectively smooths these variances, offering the most robust overall performance.

\paragraph{MME (Table~\ref{tab:mme})}
On MME, which appears to be the most challenging subset with the lowest baseline scores, the benefits of vendor diversity are most pronounced. The Mixed-Vendor MAC achieves a Recall@1 of 40.00\%, outperforming the best Single-LLM (Gemini-2.5-Pro) by 10\% and the best Single-Vendor MAC (Claude-4.5-Sonnet) by 5\%. This dominance extends consistently across all retrieval depths, with the Mixed-Vendor MAC maintaining a clear lead at Recall@3, 5, and 10. This suggests that when individual models struggle, the diverse reasoning strategies provided by a mixed panel are critical for surfacing the correct diagnosis that homogeneous teams miss.

\paragraph{LIRICAL and HMS (Tables~\ref{tab:hms} \& ~\ref{tab:lirical})}
On the LIRICAL and HMS subsets, the Mixed-Vendor MAC demonstrates high stability and top-tier performance. On \textbf{HMS}, the Mixed-Vendor MAC achieves highest overall performance across all metrics, outperforming the strongest standalone model (Gemini-2.5-pro) as well as all Single-Vendor MAC configurations. On \textbf{LIRICAL}, the Mixed-Vendor MAC achieves a Recall@1 of 35.69\%, performing competitively with the best-performing Single-LLM (Gemini-2.5-Pro at 36.24\%) while surpassing all baselines at deeper retrieval levels, including Recall@3, 5, and 10. Moreover, the Mixed-Vendor MAC maintains better performance than all single-vendor configurations as depth increases. 

Upon examining the evaluations provided by the o4-mini judge, we observed a recurring discrepancy in the HMS subset where technically correct, granular predictions were penalized. To ensure metric validity, the HMS subset ($n=88$) underwent independent clinical adjudication by a medical doctor across all experimental configurations (Single-LLM, Single-Vendor MAC, and Mixed-Vendor MAC). The review revealed that the MAC consensus process often drives agents toward "hyper-specific" etiologic diagnoses that correctly incorporate patient phenotypes not captured by the broader gold labels, resulting in initial penalties by the o4-mini judge. For instance, in Case HMS-73, the models identified \textit{Primary Sjögren’s Syndrome with lymphomatous transformation}; while the gold label was simply \textit{Primary Sjögren syndrome}, the expert confirmed the model’s refinement was clinically supported by the recorded phenotype of "Night sweats." Similar refinements occurred in cases of \textit{Cryoglobulinemic vasculitis} where agents correctly identified \textit{Hepatitis C} as the underlying driver. By manually verifying these ``hyper-specific'' cases as correct, we ensured the reported HMS scores reflect the framework's true diagnostic depth rather than being artificially deflated by automated judge sensitivity to semantic granularity.

Notably, our results reveal that scaling up to a multi-agent team of identical models does not guarantee improvement over the single-model baseline. In some cases, Single-Vendor MACs actually underperform their standalone base models, most prominently with Gemini on both LIRICAL and HMS, and Claude on HMS. This degradation likely stems from \textbf{correlated failure modes}. As noted earlier, we removed role-specialized prompts to isolate the impact of model diversity in the MAC framework. In this absence of role-induced variance, when homogenous agents discuss and deliberate, they risk reinforcing shared hallucinations or collectively drifting away from a correct initial hypothesis. The Mixed-Vendor MAC avoids this pitfall, effectively stabilizing performance where single-vendor teams falter.

\subsection{DiagnosisArena}

Table~\ref{tab:diagnosis_arena} presents the results on DiagnosisArena, a benchmark of complex real-world case reports. This dataset is notably difficult for individual models: Gemini-2.5-Pro and Claude-4.5-Sonnet struggle significantly in the Single-LLM setting, achieving only 20.00\% and 19.39\% Top-1 accuracy, respectively lagging behind the stronger Single-LLM o4-mini baseline (32.12\%). 

Yet, the Mixed-vendor MAC configuration achieves a highest Top-1 accuracy of 36.36\% and Top-5 accuracy of 49.09\%. Crucially, although two of the three agents in the mixed team perform poorly in isolation, their distinct reasoning processes allow the Mixed-Vendor system to exceed the strong o4-mini Single-Vendor MAC (35.76\%). This result demonstrates that the Mixed-Vendor framework is robust to performance disparities among its agents; it does not strictly require uniformly high-performing constituents to succeed. By integrating these diverse perspectives, the supervisor gains unique insights from the ``weaker" agents to resolve cases that a homogenous strong team would collectively miss, highlighting the systemic resilience of the Mixed-Vendor approach.

\begin{table}[h]
\centering
\small
\caption{Performance on \textbf{DiagnosisArena} (n=165). OpenAI refers to o4-mini, Gemini refers to Gemini-2.5-Pro, and Claude refers to Claude-4.5-Sonnet.}
\label{tab:diagnosis_arena}
\begin{tabular}{lrr}
\toprule
\textbf{Setup} & \textbf{Top-1} & \textbf{Top-5} \\
\midrule
Single-LLM (OpenAI) & 32.12\% & 46.06\% \\
Single-LLM (Gemini) & 20.00\% & 31.51\% \\
Single-LLM (Claude) & 19.39\% & 29.70\% \\
Single-Vendor MAC (OpenAI) & 35.76\% & 47.88\% \\
Single-Vendor MAC (Gemini) & 33.94\% & 44.24\% \\
Single-Vendor MAC (Claude) & 32.73\% & 45.45\% \\
\textbf{Mixed-Vendor MAC} & \textbf{36.36\%} & \textbf{49.09\%} \\
\bottomrule
\end{tabular}
\end{table}

\subsection{Overlap and Diversity Analysis}\label{sec:overlap_diversity}

\begin{figure*}[!t]
    \centering
    \includegraphics[width=1.6\columnwidth]{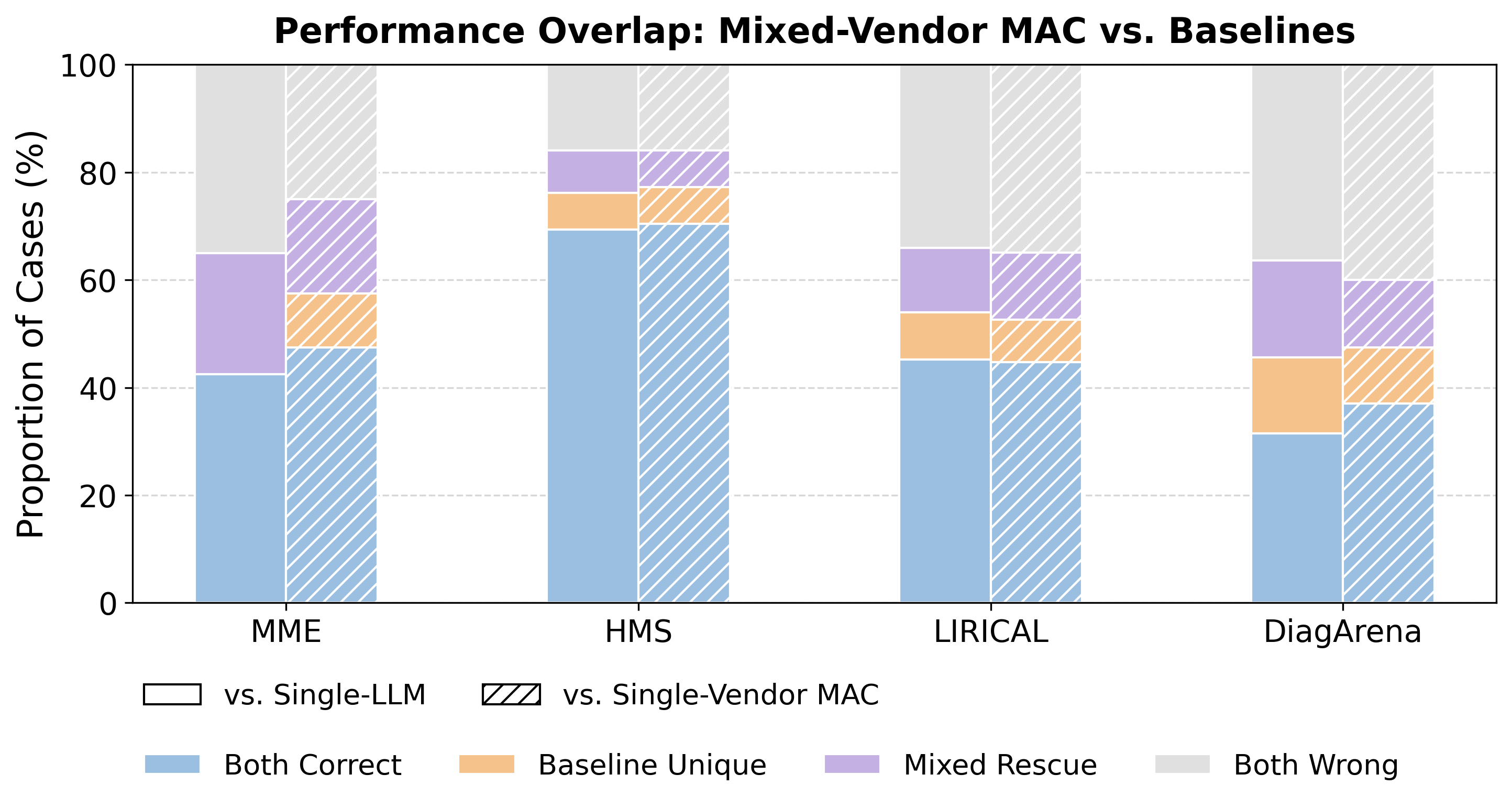}
    \caption{Analysis of Correct Prediction Overlap. The x-axis shows the dataset. For the RareBench datasets (MME, HMS, LIRICAL), overlap is calculated using \textbf{Recall@10}. For DiagnosisArena, overlap is calculated using \textbf{Top-5 Accuracy}. In each pair, the left bar compares Mixed-Vendor MAC against the Best Single LLM, and the right bar compares it against the Best Single-Vendor MAC.}
    \label{fig:figure1}
\end{figure*}

To understand the mechanism driving the performance of the Mixed-Vendor MAC, we analyze the relationship between the solution sets of different configurations. We hypothesize that the mixed-vendor systems succeed by aggregating complementary inductive biases, thereby approximating the union of individual model capabilities. By integrating diverse reasoning priors, the system should effectively expand the total valid solution space beyond the blind spots of any single vendor.

To test this, we decompose the predictions of the Mixed-Vendor MAC ($M$) relative to two baselines ($S$) for each dataset: the best-performing Single LLM and the best-performing Single-Vendor MAC. For each comparison, we categorize the solution sets into four mutually exclusive groups shown in Figure \ref{fig:figure1}: \begin{itemize}
    \item \textbf{Mutually Correct} ($S \cap M$): Cases where both systems predicted the correct diagnosis.
    \item \textbf{Baseline Unique} ($S \setminus M$): Cases where only the baseline predicted the correct diagnosis.
    \item \textbf{Mixed Rescue} ($M \setminus S$): Cases where only the Mixed system predicted the correct diagnosis.
    \item \textbf{Both Wrong}: Cases where neither system predicted the correct diagnosis.
\end{itemize}

We then quantify the overlap and asymmetry of these sets using two metrics:

\begin{itemize}
    \item \textbf{$\Delta$Coverage}: Measures the net expansion of knowledge, defined as $\text{Coverage}(S \to M) - \text{Coverage}(M \to S)$, where the base metric $\text{Coverage}(A \to B) = \frac{|A \cap B|}{|A|}$. Positive values indicate that the Mixed system ($M$) represents a net expansion of the Single system ($S$). 
    \item \textbf{Jaccard Index} ($J(S, M) = \frac{|S \cap M|}{|S \cup M|}$): Measures the overall similarity of solution sets.
\end{itemize}

Our analysis of these metrics yields three key findings: (1) mixed-vendor teams operate by "rescuing" more diagnoses than they discard during consensus; (2) performance gains are inversely correlated with model similarity, peaking when models possess distinct areas of expertise that compensate for each other's weakness; and (3) homogeneous teams remain limited by correlated failure modes that conversational debate alone cannot resolve.

\paragraph{Mixed-Vendor Consensus Rescues More Diagnoses than it Discards} 
Figure \ref{fig:figure1} visualizes the trade-off between cases gained (Mixed Rescue) versus cases lost (Baseline Unique). On all four datasets, the "Mixed Rescue" fraction consistently exceeds the "Baseline Unique" fraction. This is most evident on MME, where the system recovers 22.50\% of cases previously missed by the strongest Single LLM baseline (Mixed Rescue) while suffering zero regression (0.00\% Baseline Unique), effectively acting as a perfect superset of the baseline knowledge. 
In other datasets, a small fraction of "Baseline Unique" cases (6.82\%--14.1\%) reflects instances where consensus dynamics override correct individual priors when multiple agents converge on the same incorrect hypothesis. 
Despite this "voting down" effect, the net result remains positive, confirming that the volume of unique insights contributed by diverse vendors vastly outweighs the dilution of any single expert's prior.

\begin{figure*}[t!]
    \centering
    \begin{minipage}[b]{0.45\textwidth}
        \centering
        \includegraphics[width=\linewidth]{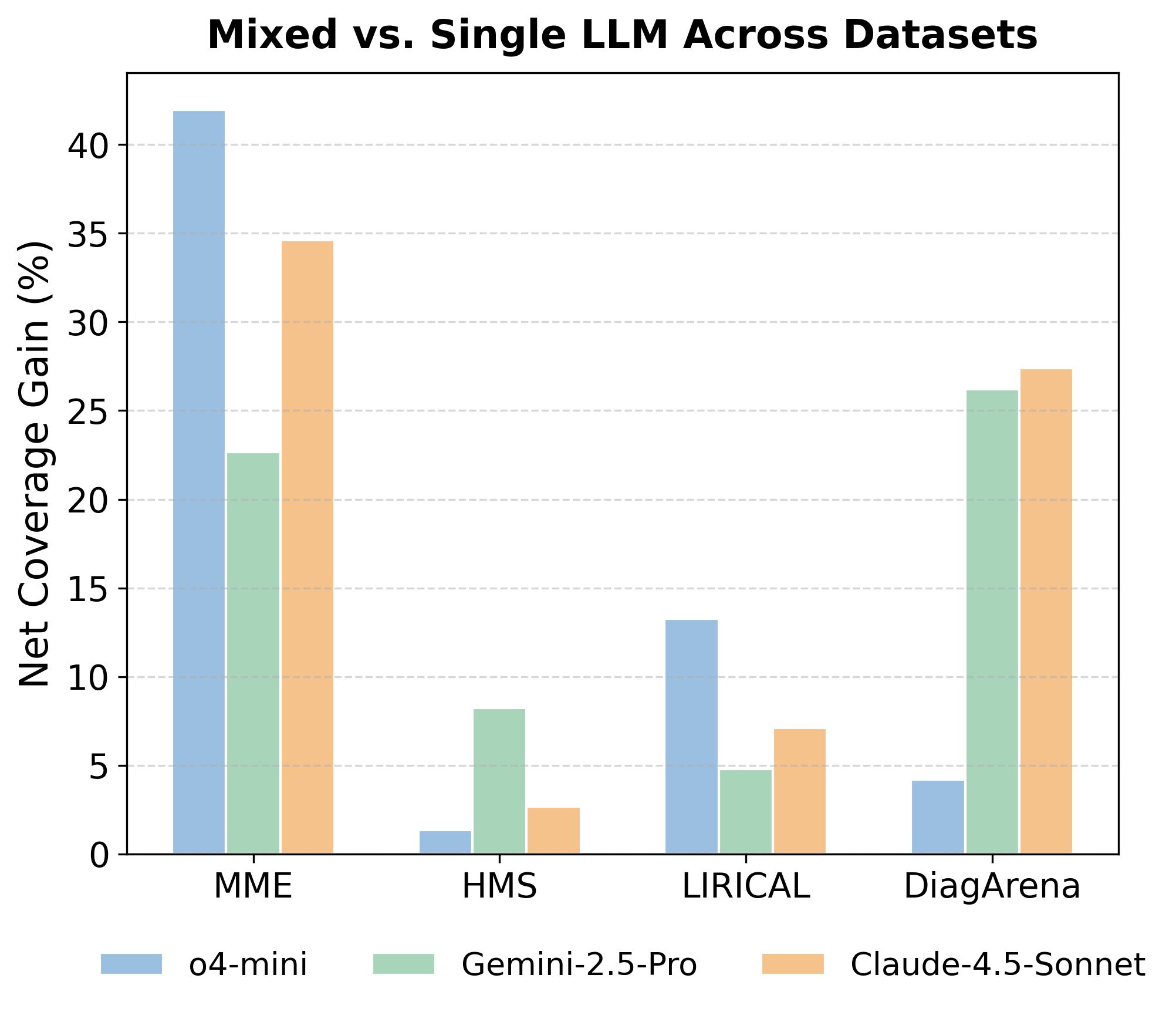}
        \caption{$\Delta\text{Coverage}$ across different datasets between Mixed-vendor MAC and Single LLMs. High positive bars indicate the Mixed model subsumes the baseline's knowledge.}
        \label{fig:figure2}
    \end{minipage}
    \hfill 
    \begin{minipage}[b]{0.53\textwidth}
        \centering
        \includegraphics[width=\linewidth]{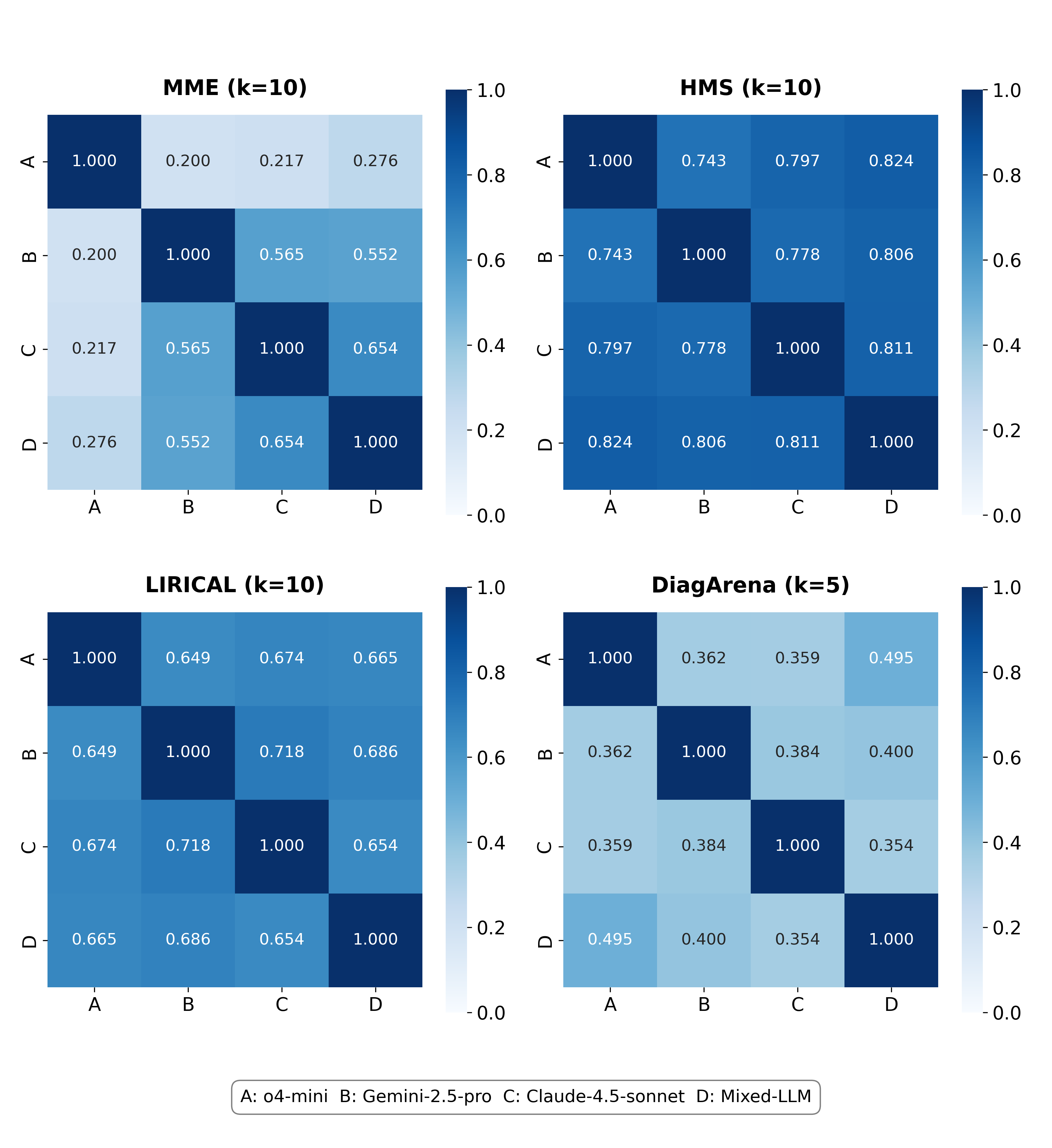}
        \caption{Pairwise Jaccard similarity heatmaps between Single LLMs and the Mixed-Vendor MAC.}
        \label{fig:heatmap_diversity}
    \end{minipage}
\end{figure*}
\begin{figure}[t!]
    \centering
    \includegraphics[width=0.9\columnwidth]{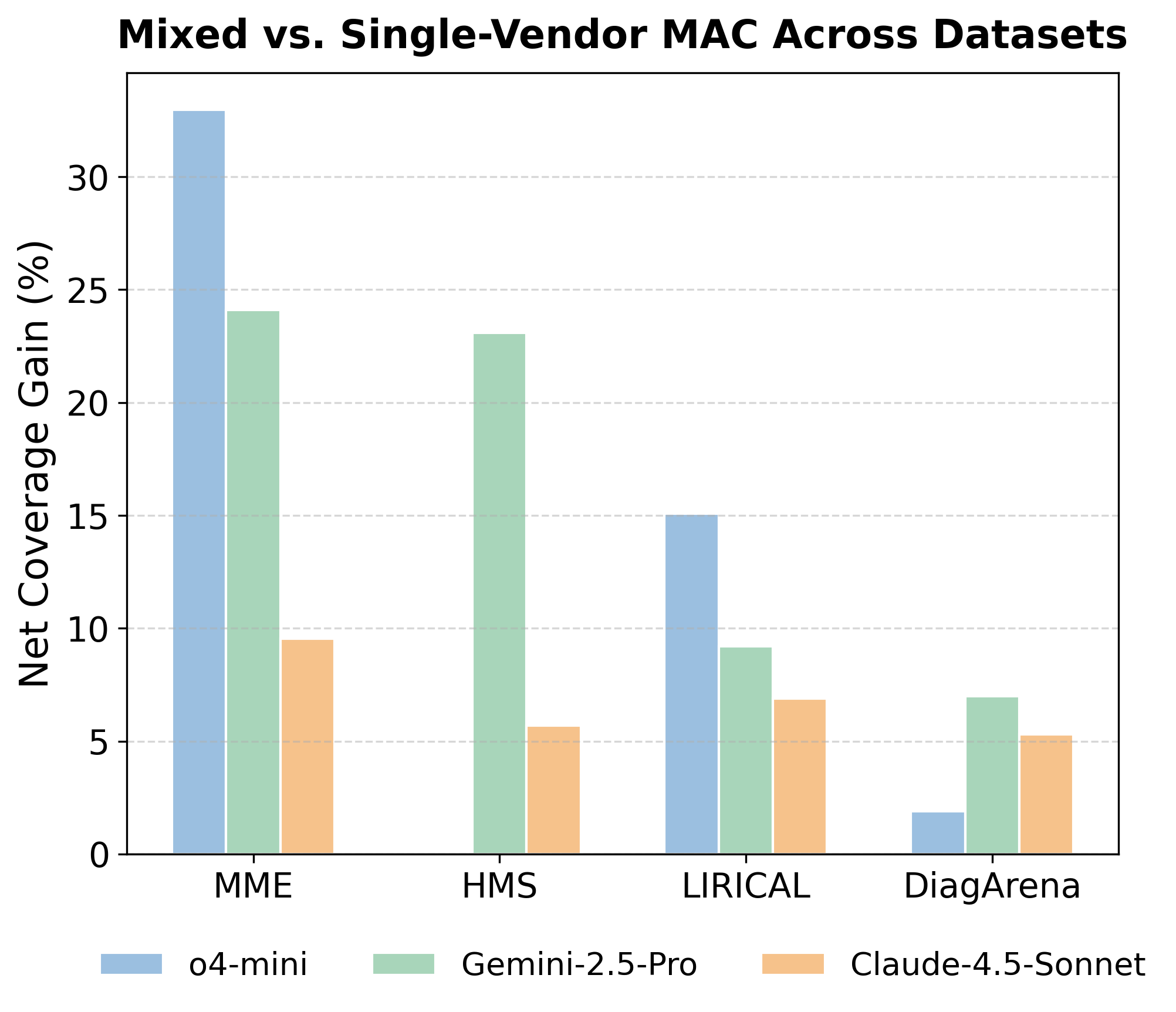}
    \caption{$\Delta\text{Coverage}$ across different datasets between Mixed-vendor MAC and Single-vendor MAC.}
    \label{fig:figure3}
\end{figure}

\paragraph{Performance Gains are Driven by Low Model Correlation (Figures \ref{fig:figure2}, \ref{fig:heatmap_diversity}, and \ref{fig:figure3})}
We analyze the relationship between solution set diversity and coverage gain. The coverage metrics in Figure \ref{fig:figure2} and the Jaccard similarities in Figure \ref{fig:heatmap_diversity} together reveal a strong inverse correlation: the Mixed system's advantage is maximized precisely when the underlying models are most distinct.

On challenging datasets like MME and DiagnosisArena, the heatmaps in Figure \ref{fig:heatmap_diversity} show low Jaccard similarity between the Single LLMs. This confirms that individual models possess orthogonal strengths, covering disjoint slices of the medical knowledge. Consequently, Figure \ref{fig:figure2} shows massive gains in $\Delta$Coverage, as the Mixed system successfully aggregates these complementary insights to envelope the capabilities of the entire field.

Conversely, high inter-model similarity on HMS (dark blue cells) suggests that individual agents capture nearly the same information. This saturation explains the low $\Delta$Coverage in Figure \ref{fig:figure2}: with little unique information to aggregate, consensus-building cannot significantly expand the solution space. However, when comparing Mixed-Vendor MAC with Single-Vendor MACs in Figure \ref{fig:figure3}, we observe an increase in average gains across HMS and LIRICAL. These gains reflect the "rescue" effect: models like Gemini suffer a performance drop when scaling from a single model to a Single-Vendor MAC on both datasets, where reinforced shared inductive biases lead teams to discard correct hypotheses originally held by the base model. The inherent diversity of the Mixed-Vendor system provides a more robust reasoning path, allowing it to effectively recover the correct diagnoses that the homogeneous teams missed. 

\paragraph{Homogeneous Teams Remain Trapped in Correlated Failure Modes.} 
Figure \ref{fig:figure3} shows that vendor diversity yields knowledge gains unattainable through homogeneous debate. Positive $\Delta$Coverage across all datasets confirms that Single-Vendor MACs remain limited by correlated failure modes that internal reasoning alone cannot resolve. The zero net gain against Single-Vendor MAC (OpenAI) on HMS occurs because the Mixed-Vendor system's ability to rescue cases missed by OpenAI is numerically offset by instances where the correct OpenAI-specific signal was suppressed by the incorrect majority (Gemini and Claude). This result shows that without sufficient cross-vendor diversity, the system cannot break past the reasoning limits of its most dominant constituent. Supplementary analysis of Single-Vendor MACs (see Appendix) confirms this mechanism: the Mixed system achieves higher gains when the competing homogeneous teams exhibit the low mutual overlap.

\subsection{Ablation Studies}
\begin{table*}[t!]
    \centering
    \small
    \setlength{\tabcolsep}{4pt} 
    
    \begin{minipage}[t]{0.48\textwidth}
        \centering
        \caption{BioLORD Evaluation on \textbf{Combined RareBench}. OpenAI refers to o4-mini, Gemini refers to Gemini-2.5-Pro, and Claude refers to Claude-4.5-Sonnet.}
        \label{tab:combined_biolord}
        \resizebox{\linewidth}{!}{
            \begin{tabular}{lrrrr}
            \toprule
            \textbf{Setup} & \textbf{R@1} & \textbf{R@3} & \textbf{R@5} & \textbf{R@10} \\
            \midrule
            Single-LLM (OpenAI) & 17.15 & 23.01 & 26.44 & 29.06 \\
            Single-LLM (Gemini) & 21.38 & 26.22 & 28.84 & 31.06 \\
            Single-LLM (Claude) & 16.34 & 23.81 & 27.04 & 31.88 \\
            Single-Vendor MAC (OpenAI) & 18.37 & 23.21 & 25.03 & 29.07 \\
            Single-Vendor MAC (Gemini) & 20.17 & 24.80 & 26.42 & 28.43 \\
            Single-Vendor MAC (Claude) & 18.35 & 25.81 & 29.25 & 32.07 \\
            \textbf{Mixed-Vendor MAC} & \textbf{22.99} & \textbf{27.42} & \textbf{30.45} & \textbf{33.68} \\
            \bottomrule
            \end{tabular}
        }
    \end{minipage}
    \hfill
    \begin{minipage}[t]{0.48\textwidth}
        \centering
        \caption{BioLORD Evaluation on \textbf{MME} subset (n=40). Model versions consistent with Table \ref{tab:combined_biolord}.}
        \label{tab:mme_biolord}
        \resizebox{\linewidth}{!}{
            \begin{tabular}{lrrrr}
            \toprule
            \textbf{Setup} & \textbf{R@1} & \textbf{R@3} & \textbf{R@5} & \textbf{R@10} \\
            \midrule
            Single-LLM (OpenAI) & 5.00 & 7.50 & 7.50 & 7.50 \\
            Single-LLM (Gemini) & 25.00 & 35.00 & 37.50 & 40.00 \\
            Single-LLM (Claude) & 2.50 & 22.50 & 30.00 & 35.00 \\
            Single-Vendor MAC (OpenAI) & 5.00 & 5.00 & 5.00 & 7.50 \\
            Single-Vendor MAC (Gemini) & 25.00 & 30.00 & 32.50 & 32.50 \\
            Single-Vendor MAC (Claude) & 27.50 & 30.00 & 35.00 & 37.50 \\
            \textbf{Mixed-Vendor MAC} & \textbf{30.00} & \textbf{35.00} & \textbf{40.00} & \textbf{42.50} \\
            \bottomrule
            \end{tabular}
        }
    \end{minipage}

    \vspace{0.6cm} 

    \begin{minipage}[t]{0.48\textwidth}
        \centering
        \caption{BioLORD Evaluation on \textbf{HMS} subset (n=88). Model versions consistent with Table \ref{tab:combined_biolord}.}
        \label{tab:hms_biolord}
        \resizebox{\linewidth}{!}{
            \begin{tabular}{lrrrr}
            \toprule
            \textbf{Setup} & \textbf{R@1} & \textbf{R@3} & \textbf{R@5} & \textbf{R@10} \\
            \midrule
            Single-LLM (OpenAI) & 42.53 & 57.47 &\textbf{67.82} & 74.71 \\
            Single-LLM (Gemini) & 47.13 & 55.17 & 63.22 & 68.97 \\
            Single-LLM (Claude) & 39.08 & 56.32 & 64.37 & 74.71 \\
            Single-Vendor MAC (OpenAI) & 47.13 & 56.32 & 63.22 & \textbf{75.86} \\
            Single-Vendor MAC (Gemini) & 43.68 & 50.57 & 56.32 & 59.77 \\
            Single-Vendor MAC (Claude) & 39.08 & 55.17 & 64.37 & 72.41 \\
            \textbf{Mixed-Vendor MAC} & \textbf{48.28} & \textbf{57.47} & 64.37 & 72.41 \\
            \bottomrule
            \end{tabular}
        }
    \end{minipage}
    \hfill
    \begin{minipage}[t]{0.48\textwidth}
        \centering
        \caption{BioLORD Evaluation on \textbf{LIRICAL} subset (n=370). Model versions consistent with Table \ref{tab:combined_biolord}.}
        \label{tab:lirical_biolord}
        \resizebox{\linewidth}{!}{
            \begin{tabular}{lrrrr}
            \toprule
            \textbf{Setup} & \textbf{R@1} & \textbf{R@3} & \textbf{R@5} & \textbf{R@10} \\
            \midrule
            Single-LLM (OpenAI) & 12.43 & 16.49 & 18.65 & 20.54 \\
            Single-LLM (Gemini) & 14.86 & 18.38 & 19.73 & 21.08 \\
            Single-LLM (Claude) & 12.43 & 16.22 & 17.84 & 21.35 \\
            Single-Vendor MAC (OpenAI) & 12.97 & 17.30 & 18.11 & 20.27 \\
            Single-Vendor MAC (Gemini) & 14.05 & 18.11 & 18.65 & 20.54 \\
            Single-Vendor MAC (Claude) & 12.43 & 18.38 & 20.27 & 21.89 \\
            \textbf{Mixed-Vendor MAC} & \textbf{16.22} & \textbf{19.46} & \textbf{21.35} & \textbf{23.51} \\
            \bottomrule
            \end{tabular}
        }
    \end{minipage}
\end{table*}
\subsubsection{Robustness to Evaluation Protocols}\label{sec:robustness_eval}
We evaluate the \textbf{RareBench} dataset using \textbf{BioLORD}, a retrieval-based protocol that maps predictions to the semantically closest standardized disease name based on cosine similarity against the OMIM \citep{hamosh2005omim}, Orphanet \citep{rath2012representation}, and CCRD \citep{ccrd2024platform}.

As shown in Table ~\ref{tab:combined_biolord}--\ref{tab:lirical_biolord}, the Mixed-Vendor system achieves the highest Recall@1 and Recall@3 across all subsets under this metric, confirming that the benefits of vendor diversity are robust.

The observed drop in absolute recall on MME and LIRICAL compared to the LLM judge is primarily due to lexical sensitivity. From qualitative analysis, we found that models often predict precise names based on genetic etiology (e.g., \textit{EXTL3-related Immunoskeletal Dysplasia}), whereas gold labels frequently use descriptive clinical names (e.g., \textit{Immunoskeletal dysplasia with neurodevelopmental abnormalities}). While the LLM judge correctly identifies these as equivalent, the embedding-based BioLORD judge penalizes the lexical distance between these names, resulting in lower absolute scores.

\subsubsection{Supervisor Model Generalization}
To verify that the observed advantages of the mixed-vendor configuration are not artifacts of a specific supervisor model, we conduct supervisor-vendor ablations on the MME dataset, the most challenging subset of RareBench while remaining compact in size. We replace the default o4-mini supervisor with Gemini-2.5-Pro or Claude-4.5-Sonnet while keeping the doctor agents constant.

\begin{table}[h!]
    \centering
    \small
    \caption{\textbf{Supervisor Ablation on MME (n=40).} Comparison of Single-Vendor and Mixed-Vendor teams under different supervisor models.}
    \label{tab:supervisor_ablation}
    \resizebox{\linewidth}{!}{
        \begin{tabular}{lrrrr}
        \toprule
        \textbf{Setup} & \textbf{R@1} & \textbf{R@3} & \textbf{R@5} & \textbf{R@10} \\
        \midrule
        \multicolumn{5}{l}{\textit{\textbf{Baselines: Single LLM}}} \\
        Single-LLM (OpenAI) & 10.00 & 22.50 & 25.00 & 27.50 \\
        Single-LLM (Gemini) & 30.00 & 37.50 & 42.50 & 47.50 \\
        Single-LLM (Claude) & 2.50 & 22.50 & 30.00 & 42.50 \\
        \midrule
        \multicolumn{5}{l}{\textit{\textbf{Supervisor: OpenAI}}} \\
        Single-Vendor MAC (OpenAI) & 15.00 & 25.00 & 30.00 & 35.00 \\
        Single-Vendor MAC (Gemini) & 35.00 & 45.00 & 47.50 & 47.50 \\
        Single-Vendor MAC (Claude) & 35.00 & 45.00 & 50.00 & 57.50 \\
        \textbf{Mixed-Vendor MAC} & \textbf{40.00} & \textbf{52.50} & \textbf{60.00} & \textbf{65.00} \\
        \midrule
        \multicolumn{5}{l}{\textit{\textbf{Supervisor: Gemini}}} \\
        Single-Vendor MAC (OpenAI) & 32.50 & 40.00 & 52.50 & 55.00 \\
        Single-Vendor MAC (Gemini) & 32.50 & 42.50 & 42.50 & 50.00 \\ 
        Single-Vendor MAC (Claude) & 32.50 & 40.00 & 40.00 & 50.00 \\
        \textbf{Mixed-Vendor MAC} & \textbf{40.00} & \textbf{50.00} & \textbf{55.00} & \textbf{57.50} \\
        \midrule
        \multicolumn{5}{l}{\textit{\textbf{Supervisor: Claude}}} \\
        Single-Vendor MAC (OpenAI) & 22.50 & 27.50 & 40.00 & 52.50 \\
        Single-Vendor MAC (Gemini) & 27.50 & 32.50 & 32.50 & 40.00 \\
        Single-Vendor MAC (Claude) & 27.50 & 37.50 & 42.50 & 45.00 \\
        \textbf{Mixed-Vendor MAC} & \textbf{42.50} & \textbf{50.00} & \textbf{52.50} & \textbf{57.50} \\
        \bottomrule
        \end{tabular}
    }
\end{table}

As shown in Table \ref{tab:supervisor_ablation}, the Mixed-Vendor configuration consistently outperforms all Single-Vendor teams, regardless of the supervisor chosen. Specifically, the mixed team achieves a Recall@1 of 40.00\%--42.50\% under Gemini and Claude supervisors, maintaining a substantial lead (+7.5\% to +15.0\%) over the best performing homogeneous baselines. This stability confirms that the performance gains stem from the diverse inductive biases of the doctor agents, rather than the specific capabilities of the aggregator.

\section{Limitations and Ethical Considerations}
\paragraph{Computational Overhead and Cost.} The MAC framework inherently introduces higher computational overhead and latency than Single-LLM baselines. However, our experiments revealed that the Mixed-Vendor configuration is not consistently the most resource-intensive setup. In several instances, Single-Vendor MAC, particularly those using Gemini or Claude, incurred higher API costs and resulted in longer conversation logs than the Mixed-Vendor MAC. A practical advantage of the Mixed-Vendor approach is that it hedges against the specific verbosity or cost spikes of a single provider, sparing researchers from having to pre-select the most efficient vendor. Nevertheless, the multi-turn nature of any MAC architecture remains a significant hurdle for time-sensitive clinical applications like emergency triage.

\paragraph{The Consensus Trap and Diagnostic Safety.} As detailed in Appendix D.2, multi-agent systems are susceptible to a ``consensus trap'' where an initially correct minority signal is progressively diluted by a cohesive but incorrect majority. While vendor diversity mitigates correlated failure modes, it cannot fully eliminate the risk of the system converging on a hallucination. In real-world clinical use, such errors could lead to misdiagnosis if the system's output is taken as absolute. To reduce potential harm, these frameworks should be deployed with specific safeguards, such as confidence flags that alert users to high levels of agent disagreement, and should always function as decision-support tools under the final adjudication of a human clinician.
\section{Conclusion}
In this work, we show that the performance gains of multi-agent diagnostic systems stem primarily from vendor diversity, rather than from scaling homogeneous agents. Across multiple benchmarks, Mixed-Vendor MACs consistently outperform both single-model baselines and Single-Vendor MACs, particularly on challenging cases where individual models exhibit complementary failure modes. Our analysis demonstrates that mixed-vendor systems expand the effective solution space by aggregating diverse inductive biases, recovering correct diagnoses missed by any single model while avoiding the correlated failures that limit homogeneous teams. These findings highlight model heterogeneity as a key design principle for robust multi-agent reasoning in high-stakes domains.

\bibliography{custom}
\clearpage
\appendix
\section{Prompt Templates}
\label{sec:appendix_prompts}

\subsection{RareBench Prompt Templates}
\label{sec:appendix_rare_prompts}

For the rare disease benchmarks (MME, HMS, LIRICAL), we utilized the following prompts to guide the multi-agent conversation toward a Top-10 differential diagnosis.\\

{\footnotesize
\noindent \textbf{Doctor Agent System Prompt}
\begin{quote}
You are \{doctor\_name\}. You are a specialist in the field of rare diseases. You will be provided and asked about a complicated clinical case; read it carefully and then provide a diverse and comprehensive differential diagnosis. EVERY TIME you speak, include your current top-10 list at the END of the message.
\end{quote}

\noindent \textbf{Medical Supervisor System Prompt}
\begin{quote}
You are the Medical Supervisor in a hypothetical scenario. You are a specialist in the field of rare diseases. You will be provided and asked about a complicated clinical case; read it carefully and then provide a diverse and comprehensive differential diagnosis.

\textbf{Your role (general):} 1. Oversee and evaluate suggestions and decisions made by the doctors. 2. Challenge diagnoses where needed. 3. Facilitate discussion and drive consensus about diagnoses ONLY.

\textbf{Strict prohibitions:} Do NOT propose or list diagnostic tests, lab work, imaging, procedures, or any workup. Do NOT discuss management, treatment, or prognosis. Do NOT include any sections titled ``SUGGESTED TESTS'', ``TESTS'', ``WORKUP'', or similar.

\textbf{Guidelines:} Promote discussion unless there's absolute consensus. Continue dialogue if any disagreement exists. Output ``TERMINATE'' only when: 1. All doctors fully agree. 2. No further discussion is needed. 3. All diagnostic possibilities are explored.

\textbf{Finalization format:} When you decide to terminate, reply ONLY with a numbered list of exactly 10 diagnoses: 1. ... 10. ... Then append the token TERMINATE on a new line.
\end{quote}

\noindent \textbf{Initial Opening Prompt}
\begin{quote}
Patient's phenotype: \{phenostr\} \\
Enumerate the top 10 most likely diagnoses. Be precise, listing one diagnosis per line, and try to cover many unique possibilities (at least 10). Let us think step by step, you must think more steps. The top 10 diagnoses are:
\end{quote}
}

\subsection{DiagnosisArena Prompts}
{\footnotesize
\noindent \textbf{Doctor System Prompt}
\begin{quote}
You are \{doctor\_name\}. You are a medical expert in clinical diagnosis. 
\textbf{Role:} 1) Analyze the patient's presentation carefully. 
\textbf{Responsibilities:} 1. Analyze case info and others' input. 2. Offer expertise-based insights. 3. Engage in discussion. 4. Critique opinions with arguments. 5. Refine approach iteratively. 
\textbf{Guidelines:} Present analysis concisely; support with reasoning; stay open to adjustments. Include top-5 list at the END of every message.
\end{quote}

\noindent \textbf{Supervisor System Prompt}
\begin{quote}
You are the Medical Supervisor. Oversee doctors and drive consensus on TOP-5 diagnoses. 
\textbf{Tasks:} 1. Evaluate suggestions. 2. Challenge missed points. 3. Facilitate discussion. 4. Drive consensus. 
\textbf{Termination Rules:} Output ``TERMINATE'' only when doctors agree and possibilities are explored. Finalization format: numbered list of exactly 5 diagnoses (1--5), then TERMINATE.
\end{quote}

\noindent \textbf{Initial Opening Prompt}
\begin{quote}
As a medical expert, please make a diagnosis based on the case info, physical exam, and diagnostic tests. Enumerate the top 5 most likely diagnoses in order. \\
\textbf{Case Info:} \{case\_text\} \\
\textbf{Physical Exam:} \{exam\} \\
\textbf{Diagnostic Tests:} \{tests\}
\end{quote}
}

\subsection{Evaluation Judge Prompt Templates}
\label{sec:appendix_eval_prompts}

For our primary evaluation, we utilized o4-mini as an automated judge to compare the predicted differential diagnoses against the reference standard. The following prompts were used to ensure the judge provided a standardized numerical rank for Recall calculation. \\

{\footnotesize
\noindent \textbf{Evaluator Evaluation Prompt}
\begin{quote}
I will now provide ten predicted diseases. If the standard diagnosis is among the predicted diseases, please output the numerical rank; otherwise, output ``No''. Output only ``No'' or a number from 1--10. If the standard diagnosis matches multiple predicted conditions, only output the top rank. Output only ``No'' or one number, with no additional output. \\ \\
\textbf{Predicted diseases:} \{predict\_diagnosis\} \\
\textbf{Standard diagnosis:} \{golden\_diagnosis\}
\end{quote}
}

\section{Model and Implementation Details}

\subsection{API Configurations}

All experiments were conducted between September and December 2025. We utilized the following specific model versions:

\begin{itemize}

    \item \textbf{OpenAI:} \texttt{o4-mini-0416}

    \item \textbf{Google:} \texttt{gemini-2.5-pro}

    \item \textbf{Anthropic:} \texttt{claude-4.5-sonnet}

\end{itemize}

All models were set to a temperature of $T=1.0$ to allow for diverse reasoning paths and to reflect the natural variability of clinical heuristics within the multi-agent debate.

\subsection{BioLORD Retrieval Details}

For the retrieval-based evaluation, we used the \texttt{BioLORD-2023} model. Predicted diagnosis strings were embedded and compared against the target ontology using cosine similarity. A match was recorded if the ground-truth diagnosis appeared within the top-$k$ closest semantic neighbors of the predicted string.

\section{Supplementary Diversity Analysis} \label{sec:appendix_diversity}
\begin{figure}[h!] \centering \includegraphics[width=1.05\linewidth]{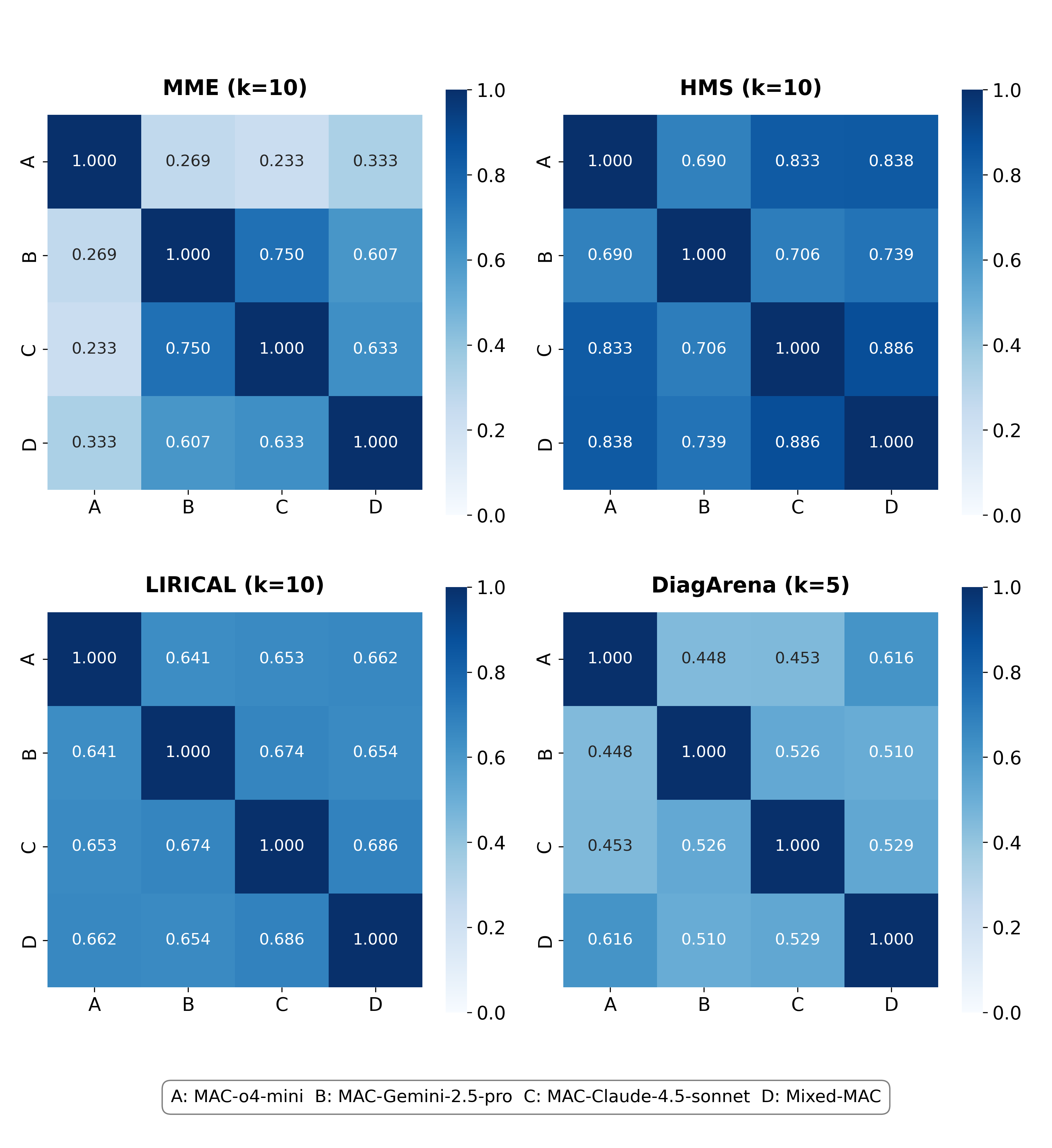} \caption{Pairwise Jaccard similarity heatmaps between Single-Vendor MAC teams (A: o4-mini, B: Gemini-2.5-pro, C: Claude-4.5-sonnet) and the Mixed-Vendor MAC (D) across benchmarks.} \label{fig:heatmap_appendix} \end{figure}

To further investigate the ``correlation trap'' discussed in Section 5.3, we present pairwise Jaccard similarity heatmaps comparing the three Single-Vendor MAC configurations against the Mixed-Vendor MAC (Figure~\ref{fig:heatmap_appendix}).

Interestingly, the Jaccard similarities between vendor-specific MAC teams are generally higher than those observed between their single-model counterparts. For example, in the MME subset, the Jaccard similarity between models B and C increases from $J=0.565$ in the single-model setting to $J=0.750$ when utilizing the MAC framework. This increased similarity likely stems from the ability of single-vendor MAC frameworks to cover a broader set of correct solutions, causing their outputs to converge. 

Furthermore, the Mixed-Vendor MAC (D) consistently maintains a relatively high Jaccard similarity with each individual constituent vendor. This synthesis effect indicates that the mixed team successfully acts as an effective knowledge bridge, aggregating disjoint slices of clinical reasoning from diverse model families into a robust final consensus.

Finally, the relationship between diversity and performance gain is most evident in the MME dataset. As illustrated in Figure~\ref{fig:figure3}, the $\Delta$Coverage peaks (32.97\%) precisely where the cross-vendor Jaccard indices are at their lowest ($J_{AB}=0.269, J_{AC}=0.233, J_{BC}=0.750$). These values indicates that the performance gain is primarily driven by the unique inductive biases of OpenAI (A) relative to the more closely correlated Gemini-Claude (B-C) pair. This inverse correlation proves that the diagnostic system derives its most significant benefits from the aggregation of uncorrelated hypotheses.

\section{Qualitative Analysis: The Mechanics of Vendor Diversity}

\label{appendix:qualitative_analysis}

In this section, we provide round-by-round diagnostic traces to illustrate how vendor diversity influences the collective reasoning of multi-agent systems. By comparing mixed-vendor teams against individual baselines and single-vendor configurations, we analyze two distinct phenomena: the ability of diverse teams to recover correct diagnoses overlooked by dominant vendors, and the risk of a correct minority signal being suppressed by a cohesive but incorrect majority consensus. These cases demonstrate that while diversity provides uncorrelated reasoning paths, the final outcome depends on the system's ability to distinguish between valid corrective signals and collective noise.

\subsection{Case Narrative: MME-14 (Cerebrocostomandibular Syndrome)}
\begin{table*}[h!]
\centering
\small
\caption{\textbf{Diagnostic Trajectory for Case MME-14.} Trajectory $(R_0 \to \dots \to R_5)$ denotes the rank of the gold label in each discussion round; $0$ indicates a miss. For Mixed-Vendor, brackets indicate: (D1: OpenAI, D2: Gemini, D3: Claude).}
\label{tab:case_study_mme14}
\begin{tabular}{p{2.8cm} c p{3.8cm} p{6.5cm}} 
\toprule
\textbf{Configuration} & \textbf{Final} & \textbf{Trajectory} & \textbf{Final Top-3 Predictions} \\
& \textbf{Rank} & & \\
\midrule
Single-LLM (OpenAI) & \textit{Missed} & N/A & 1. Isolated PRS \\
& & & 2. Stickler Syndrome \\
& & & 3. 22q11.2 Deletion \\
\midrule
Single-LLM (Gemini) & 7 & N/A & 1. Stickler Syndrome \\
& & & 2. Campomelic Dysplasia \\
& & & 3. 22q11.2 Deletion \\
\midrule
Single-LLM (Claude) & 7 & N/A & 1. Stickler Syndrome \\
& & & 2. 22q11.2 Deletion \\
& & & 3. Treacher Collins \\
\midrule
Single-Vendor MAC (OpenAI) & \textit{Missed} & $0 \to 0 \to 0 \to 0 \to 0 \to 0$ & 1. Isolated PRS \\
& & & 2. Stickler Syndrome \\
& & & 3. 22q11.2 Deletion \\
\midrule
Single-Vendor MAC (Gemini) & 3 & $6 \to 4 \to 3 \to 3 \to 3 \to 3$ & 1. Campomelic Dysplasia \\
& & & 2. SEDC \\
& & & \textbf{3. CCMS} \\
\midrule
Single-Vendor MAC (Claude) & 6 & $6 \to 2 \to 2 \to 6 \to 6 \to 6$ & 1. Stickler Syndrome \\
& & & 2. Campomelic Dysplasia \\
& & & 3. Nager Syndrome \\
\midrule
\textbf{Mixed-Vendor} & \textbf{1} & $0\text{(D1)} \to 3\text{(D2)} \to 3\text{(D3)} \to$ & \textbf{1. CCMS} \\
\textbf{MAC} & & $1\text{(D1)} \to 1\text{(D2)} \to 1\text{(D3)}$ & 2. Spondylocostal Dysostosis \\
& & & 3. Spondylothoracic Dysostosis \\
\bottomrule
\end{tabular}
\end{table*}

We analyze Case MME-14 from the RareBench MME subset.  This case serves as an example of successful recovery, where the mixed-vendor system identifies a correct diagnosis that was missed by a single-vendor team. 

\paragraph{Case Narrative:} 
Patient presents with Pierre-Robin sequence (micrognathia, glossoptosis, and cleft palate), a high-arched palate, and abnormal thorax morphology (specifically rib gaps/defects).

\paragraph{Gold Diagnosis}: Cerebrocostomandibular Syndrome (CCMS)

\paragraph{Discussion:}
As shown in Table~\ref{tab:case_study_mme14}, the \textbf{Single-Vendor MAC (OpenAI)} remains trapped in an ``echo chamber,'' where multiple rounds of internal reasoning fail to resolve a shared blind spot regarding the thoracic cues. While the base OpenAI model prioritizes common syndromes like Stickler or 22q11.2 deletion, the homogeneous team reinforces this bias rather than correcting it.

In contrast, the \textbf{Mixed-Vendor MAC} demonstrates a successful rescue. Although the first agent (OpenAI) initially misses the diagnosis in Round 0, the introduction of \textbf{CCMS} by the second agent (Gemini) in Round 1 provides a critical corrective signal. By Round 3, the diverse reasoning paths allow the entire team---including the previously biased OpenAI agent---to converge on the correct diagnosis at \textbf{Rank 1}. This confirms that vendor diversity provides a robust reasoning path capable of amplifying correct minority signals that are otherwise discarded by homogeneous teams.

\subsection{Case Narrative: HMS-12 (The Consensus Trap)}
\label{appendix:case_study_hms12}

\begin{table*}[h!]
\centering
\small
\caption{\textbf{Temporal Diagnostic Trajectory for Case HMS-12.} Trajectory $(R_0 \to \dots \to R_5)$ denotes the rank of the gold label in each discussion round; $0$ indicates a miss. For Mixed-Vendor, brackets indicate: (D1: OpenAI, D2: Gemini, D3: Claude).}
\label{tab:case_study_hms12}
\begin{tabular}{p{2.8cm} c p{3.8cm} p{6.5cm}} 
\toprule
\textbf{Configuration} & \textbf{Final} & \textbf{Trajectory} & \textbf{Final Top-3 Predictions} \\
& \textbf{Rank} & & \\
\midrule
Single-LLM (OpenAI) & 1 & N/A & 1. \textbf{Primary FSGS} \\
& & & 2. Obesity-related glomerulopathy \\
& & & 3. Hypertensive nephrosclerosis \\
\midrule
Single-LLM (Gemini) & \textit{Missed} & N/A & 1. Glycogen Storage Disease Type I \\
& & & 2. PRPP Synthetase Superactivity \\
& & & 3. Systemic Lupus Erythematosus \\
\midrule
Single-LLM (Claude) & \textit{Missed} & N/A & 1. Cushing Syndrome \\
& & & 2. Metabolic Syndrome \\
& & & 3. Familial Partial Lipodystrophy \\
\midrule
Single-Vendor MAC (OpenAI) & \textbf{1} & $1 \to 1 \to 1 \to 1 \to 1 \to 1$ & \textbf{1. Primary FSGS} \\
& & & 2. Secondary FSGS (Obesity) \\
& & & 3. Cushing's Syndrome \\
\midrule
Single-Vendor MAC (Gemini) & \textit{Missed} & $4 \to 0 \to 0 \to 0 \to 0 \to 0$ & 1. Glycogen Storage Disease Type I \\
& & & 2. Uromodulin-Associated Kidney \\
& & & 3. Sarcoidosis \\
\midrule
Single-Vendor MAC (Claude) & \textit{Missed} & $0 \to 6 \to 0 \to 0 \to 0 \to 0$ & 1. Cushing's Syndrome \\
& & & 2. Obesity-related glomerulopathy \\
& & & 3. Familial Partial Lipodystrophy \\
\midrule
\textbf{Mixed-Vendor} & \textit{Missed} & $1\text{(D1)} \to 6\text{(D2)} \to 0\text{(D3)} \to$ & 1. Familial Partial Lipodystrophy \\
\textbf{MAC} & & $0\text{(D1)} \to 0\text{(D2)} \to 0\text{(D3)}$ & 2. Acquired Generalized Lipodystrophy \\
& & & 3. Endogenous Cushing’s Syndrome \\
\bottomrule
\end{tabular}
\end{table*}

While vendor diversity often facilitates the rescue of rare diagnoses, it can also introduce a ``consensus trap'' where a correct minority signal is discarded in favor of a cohesive but incorrect majority. We analyze Case HMS-12 from the HMS dataset.

\paragraph{Case Narrative:} 
A complex multi-system presentation including Striae distensae (skin stretch marks), Hyperuricemia (gout-related markers), Hypertriglyceridemia, and significant renal findings: microscopic hematuria, Nephrosclerosis, and a decreased glomerular filtration rate.

\paragraph{Gold Diagnosis:} Focal Segmental Glomerulosclerosis (FSGS)

\paragraph{Discussion of Dynamics:}
As presented in Table~\ref{tab:case_study_hms12}, HMS-12 illustrates a scenario where vendor diversity leads to the suppression of a correct hypothesis. Individually, the OpenAI agent (D1) correctly identifies the gold label at Rank 1. However, in the Mixed-Vendor setting, the Gemini (D2) and Claude (D3) agents are both strongly biased toward metabolic and endocrine etiologies (Lipodystrophy and Cushing's), which also explain the patient's striae and dyslipidemia.

As the trajectory reveals, the correct signal from D1 in Round 0 is progressively diluted. By Round 2, the team has reached a cohesive but incorrect consensus on Lipodystrophy. Even the OpenAI agent (D1), which was correct in the Single-Vendor setting, is eventually swayed by the confident reasoning of its peers in Round 3, leading to a final consensus that misses the gold label entirely. This confirms that while diversity provides uncorrelated hypotheses, it requires a supervisor capable of protecting minority signals from being overwhelmed by majority-driven hallucinations.

\section{Representative Diagnostic Transcripts}
\label{appendix:transcripts}
In this section, we provide a representative diagnostic trace to illustrate the internal reasoning and collaborative dynamics of the Mixed-Vendor MAC framework. By detailing the round-by-round interaction for Case MME-14, we demonstrate the specific mechanics of a "successful recovery"—where the pooling of diverse inductive biases allows the team to surface a correct diagnosis missed by single-vendor baselines.

\vspace{0.2cm} \noindent \textbf{Case Narrative:} Patient presents with Pierre-Robin sequence (micrognathia, glossoptosis, and cleft palate), a high-arched palate, and abnormal thorax morphology (specifically rib gaps/defects).\\
\textbf{Gold Diagnosis:} Cerebrocostomandibular Syndrome (CCMS)

\footnotesize
\begin{description}
    \item[Supervisor:] Patient's phenotype: Pierre-Robin sequence, High palate, Abnormal thorax morphology. Enumerate the top 10 most likely diagnoses. Be precise, listing one diagnosis per line, and try to cover many unique possibilities (at least 10). Let us think step by step, you must think more steps. The top 10 diagnoses are:

    \item[Doctor 1:] Let's break down the key features and then map them to known syndromes or disorders:
    \begin{enumerate}
        \item Pierre-Robin sequence (micrognathia, glossoptosis, cleft palate) is a grouping of anomalies that can be isolated or part of multiple genetic syndromes.
        \item A high (arched) palate typically reflects an underlying cleft-palate spectrum or connective-tissue involvement.
        \item An abnormal thoracic morphology (narrow/bell-shaped thorax, chest wall dysplasia) points toward skeletal dysplasias or segmentation defects.
    \end{enumerate}

    Based on these three pillars, here is a diverse top-10 differential:
    \begin{enumerate}
        \item Stickler syndrome (COL2A1/COL11A1): connective-tissue disorder with Pierre-Robin and midface hypoplasia
        \item 22q11.2 deletion syndrome (DiGeorge/Velocardiofacial): cleft palate, micrognathia, occasional chest wall anomalies
        \item Treacher Collins syndrome (TCOF1): mandibulofacial dysostosis with micrognathia and palate involvement
        \item Nager syndrome (SF3B4-related acrofacial dysostosis): Pierre-Robin plus preaxial limb defects
        \item Campomelic dysplasia (SOX9): severe bent long bones, bell-shaped thorax, Robin sequence in some cases
        \item Spondylothoracic dysostosis / Jarcho-Levin syndrome (MESP2): rib-vertebral segmentation defect, ``crab-like'' thorax
        \item Kabuki syndrome (KMT2D/KDM6A): distinctive face, high-arched palate, skeletal anomalies including pectus
        \item Isolated Pierre-Robin sequence: non-syndromic occurrence of micrognathia–glossoptosis–cleft palate
        \item CHARGE syndrome (CHD7): coloboma, heart defects, choanal atresia, cleft palate, occasional chest wall findings
        \item Ellis-van Creveld syndrome (EVC/EVC2): short ribs, narrow thorax, postaxial polydactyly, occasional clefting
    \end{enumerate}

    \textbf{Top-10 current differential:} \\
    1. Stickler syndrome, 2. 22q11.2 deletion syndrome, 3. Treacher Collins syndrome, 4. Nager syndrome, 5. Campomelic dysplasia, 6. Spondylothoracic dysostosis (Jarcho-Levin), 7. Kabuki syndrome, 8. Isolated Pierre-Robin sequence, 9. CHARGE syndrome, 10. Ellis-van Creveld syndrome.

    \item[Doctor 2:] An excellent and comprehensive initial list, Doctor1. You've covered the major categories well, from common microdeletions to classic connective tissue disorders and skeletal dysplasias.
    
    I'd like to build on that by focusing even more tightly on the powerful syndromic link between Pierre-Robin sequence and intrinsic skeletal or rib cage abnormalities. The ``abnormal thorax morphology'' is a highly specific clue that, in my opinion, significantly raises the index of suspicion for certain rare disorders where this combination is a cornerstone feature.

    I would propose adding a few specific diagnoses that fit this triad almost perfectly:
    \begin{itemize}
        \item \textbf{Cerebrocostomandibular syndrome (CCMS):} This is a rare but classic triad of Pierre-Robin sequence, posterior rib gap defects (creating a very specific thoracic appearance on X-ray), and severe intellectual disability. The thoracic abnormality is pathognomonic.
        \item \textbf{Otopalatodigital (OPD) spectrum disorders:} Specifically, the more severe forms like Melnick-Needles syndrome can present with Robin sequence, a small thorax with characteristic thin or ``ribbon-like'' ribs, and other skeletal anomalies. This is an X-linked condition caused by mutations in \textit{FLNA}.
        \item \textbf{Fetal Akinesia Deformation Sequence (FADS):} It's also worth considering a different pathogenic mechanism. If the primary issue is reduced fetal movement (akinesia), this can lead to a secondary deformation sequence including micrognathia, cleft palate, joint contractures, and a small, bell-shaped thorax from pulmonary hypoplasia.
    \end{itemize}

    By incorporating these, we can refine the differential to prioritize conditions where the thoracic and craniofacial findings are known to be mechanistically linked.

    My current top-10 list, ordered by what I consider the most specific fits first:
    \begin{enumerate}
        \item Stickler syndrome (COL2A1/COL11A1/COL11A2)
        \item Campomelic dysplasia (SOX9)
        \item Cerebrocostomandibular syndrome (SNRPB)
        \item Spondylocostal/Spondylothoracic Dysostosis (Jarcho-Levin) (DLL3/MESP2/LFNG)
        \item Otopalatodigital Spectrum Disorder (e.g., Melnick-Needles) (FLNA)
        \item 22q11.2 deletion syndrome
        \item Larsen syndrome (FLNB)
        \item Nager syndrome (SF3B4)
        \item Fetal Akinesia Deformation Sequence (e.g., Pena-Shokeir syndrome)
        \item Isolated Pierre-Robin sequence
    \end{enumerate}

    \item[Doctor 3:] Excellent analysis by both colleagues. Doctor1 provided a solid foundation, and Doctor2 has astutely zeroed in on conditions where the thoracic abnormality is a \textbf{defining} feature rather than an occasional finding. As a rare disease specialist, I want to emphasize additional mechanistic considerations and rare entities that might be overlooked.

    Let me add several critical points:
    \begin{enumerate}
        \item \textbf{Diastrophic dysplasia (SLC26A2):} This sulfate transporter defect causes severe skeletal dysplasia with characteristic ``hitchhiker thumb,'' cleft palate (often Pierre-Robin), and progressive kyphoscoliosis with thoracic cage restriction. The thoracic involvement is progressive and highly significant.
        \item \textbf{Weissenbacher-Zweymüller syndrome:} An allelic variant of otospondylomegaepiphyseal dysplasia (OSMED), caused by COL11A2 mutations. Features Pierre-Robin, rhizomelic limb shortening, and dumbbell-shaped long bones with potential thoracic narrowing.
        \item \textbf{Multiple pterygium syndromes (Escobar/Lethal forms):} These can present with Robin sequence secondary to decreased fetal movement, along with webbing and a restrictive thorax from pulmonary hypoplasia.
        \item \textbf{Achondrogenesis type II / Hypochondrogenesis:} Severe COL2A1 spectrum disorder with marked thoracic hypoplasia, though often lethal or extremely severe.
        \item \textbf{Toriello-Carey syndrome:} Rare but includes Robin sequence, agenesis of corpus callosum, and thoracic/cardiac anomalies.
    \end{enumerate}

    The key discriminating factors will be:
    \begin{itemize}
        \item \textbf{Type} of thoracic abnormality (narrow/bell-shaped vs. rib gaps vs. pectus deformity)
        \item Presence of limb anomalies
        \item Vertebral segmentation defects
        \item Cardiac involvement
        \item Neurological features
    \end{itemize}

    \textbf{My current top-10 differential:}
    \begin{enumerate}
        \item Stickler syndrome (COL2A1/COL11A1/COL11A2)
        \item Campomelic dysplasia (SOX9)
        \item Cerebrocostomandibular syndrome (SNRPB)
        \item Spondylocostal dysostosis (DLL3/MESP2/LFNG/HES7)
        \item Diastrophic dysplasia (SLC26A2)
        \item 22q11.2 deletion syndrome
        \item Otopalatodigital spectrum disorders (FLNA)
        \item Weissenbacher-Zweymüller syndrome / OSMED (COL11A2)
        \item Nager syndrome (SF3B4)
        \item Larsen syndrome (FLNB)
    \end{enumerate}

    \item[Supervisor:] Colleagues, we have a rich pool of candidates but need to converge on our top-10, prioritizing the triad's most specific syndromic associations. Here are the main areas of divergence:
    \begin{enumerate}
        \item Doctor1 included syndromes with only occasional thoracic involvement (Treacher Collins, Kabuki, CHARGE).
        \item Doctor2/3 championed entities where abnormal thoracic morphology is a defining feature (Cerebrocostomandibular syndrome, Otopalatodigital/Melnick-Needles, Diastrophic dysplasia, Weissenbacher-Zweymüller, spondylocostal dysostosis).
        \item There remains debate over whether to include more common but less specifically thoracic-centered diagnoses such as 22q11.2 deletion syndrome, Nager syndrome, Stickler syndrome, fetal akinesia sequence, or Ellis-van Creveld.
    \end{enumerate}

    To drive consensus, let's rank by the strength of thoracic pathology as a core feature:
    \begin{itemize}
        \item \textbf{Pathognomonic thoracic defect:}
        \begin{itemize}
            \item Cerebrocostomandibular syndrome
            \item Spondylocostal (Jarcho-Levin) dysostosis
            \item Otopalatodigital spectrum (Melnick-Needles)
        \end{itemize}
        \item \textbf{Prominent thoracic involvement but broader skeletal phenotype:}
        \begin{itemize}
            \item Campomelic dysplasia
            \item Diastrophic dysplasia
            \item Weissenbacher-Zweymüller syndrome
            \item Larsen syndrome
        \end{itemize}
        \item \textbf{Frequently associated with PR sequence but less specific thorax signs:}
        \begin{itemize}
            \item Stickler syndrome
            \item Nager syndrome
            \item 22q11.2 deletion syndrome
            \item Fetal akinesia deformation sequence
            \item Ellis-van Creveld syndrome
        \end{itemize}
    \end{itemize}

    Questions for refining our final list:
    \begin{enumerate}
        \item Which three ``less specific thorax'' conditions nevertheless deserve inclusion in the top-10 based on frequency or clinical importance?
        \item Should we drop any of the seven ``prominent thoracic'' disorders in favor of a broader syndrome like EVC or FADS?
        \item Are there strong arguments to retain or remove Nager or 22q11.2 in this context?
    \end{enumerate}
    Please weigh in so we can finalize exactly ten.

    \item[Doctor 1:] Step-by-step reasoning toward our final top-10:
    \begin{enumerate}
        \item Pathognomonic thoracic defects must be prioritized.
        \begin{itemize}
            \item Cerebrocostomandibular syndrome (CCMS)
            \item Spondylocostal dysostosis (Jarcho-Levin)
            \item Otopalatodigital spectrum (Melnick-Needles)
        \end{itemize}
        \item Equally, we need the major generalized skeletal dysplasias with a consistently narrow/bell-shaped thorax plus Robin sequence:
        \begin{itemize}
            \item Campomelic dysplasia
            \item Diastrophic dysplasia
            \item Weissenbacher-Zweymüller syndrome
            \item Larsen syndrome
        \end{itemize}
        \item Finally, three of the ``less thorax-specific but clinically common'' syndromes are essential due to their prevalence and impact on management:
        \begin{itemize}
            \item Stickler syndrome
            \item 22q11.2 deletion syndrome
            \item Nager syndrome
        \end{itemize}
    \end{enumerate}
    We debated whether to drop Larsen in favor of Ellis-van Creveld or Fetal Akinesia Deformation Sequence, but Larsen's consistent rib/vertebral involvement in infancy tips the balance in its favor for our core ten.

    \textbf{Final Consensus Top-10 Differential Diagnosis:}
    \begin{enumerate}
        \item Cerebrocostomandibular syndrome (SNRPB)
        \item Spondylocostal dysostosis / Jarcho-Levin syndrome (DLL3, MESP2, LFNG, etc.)
        \item Melnick-Needles syndrome (FLNA; severe form of otopalatodigital spectrum)
        \item Campomelic dysplasia (SOX9)
        \item Diastrophic dysplasia (SLC26A2)
        \item Weissenbacher-Zweymüller syndrome / OSMED (COL11A2)
        \item Larsen syndrome (FLNB)
        \item Stickler syndrome (COL2A1, COL11A1, COL2A1/COL11A1/COL11A2)
        \item 22q11.2 deletion syndrome
        \item Nager syndrome (SF3B4)
    \end{enumerate}

    \item[Doctor 2:] This is a pivotal finding. The description of ``multiple posterior rib gaps'' and a ``'crab-like' thoracic cage'' is not just a general descriptor of an abnormal thorax; it is a highly specific, almost pathognomonic radiological sign that dramatically narrows our differential. This information allows us to move from a broad list of possibilities to a very focused set of top candidates.
    
    My thoughts on this are as follows:
    \begin{enumerate}
        \item \textbf{The Significance of ``Posterior Rib Gaps'':} This is the hallmark feature of \textbf{Cerebrocostomandibular syndrome (CCMS)}. In this condition, the posterior portions of the ribs fail to form, leaving cartilaginous or fibrous bands that are radiolucent on X-ray. The combination of these specific rib defects with Pierre-Robin sequence is the classic triad for CCMS. This diagnosis, which was a strong consideration before, now moves to the very top of the list.
        \item \textbf{The Significance of ``'Crab-like' Thoracic Cage'':} This descriptor is most famously associated with \textbf{Spondylocostal Dysostosis (SCD) / Jarcho-Levin syndrome}. In SCD, extensive vertebral segmentation defects (hemivertebrae, fused vertebrae) cause the ribs to emerge from the spine in a fan-like or ``crab-like'' array. While the primary defect is vertebral, the thoracic appearance is profound. Therefore, SCD remains an extremely high-yield diagnosis. The key differentiator will be a spine X-ray: if it shows widespread vertebral anomalies, SCD is favored; if the spine is relatively normal but the ribs have intrinsic gaps, CCMS is the diagnosis.
        \item \textbf{Impact on the Previous Differential:} This finding effectively rules out or makes extremely unlikely most of the other diagnoses we were considering.
        \begin{itemize}
            \item \textbf{Stickler, Larsen, and Nager syndromes} do not characteristically feature posterior rib gaps or a crab-like thorax.
            \item \textbf{Campomelic and Diastrophic dysplasia} cause a narrow, bell-shaped thorax, but not due to specific rib gaps.
            \item \textbf{Otopalatodigital spectrum disorders} (like Melnick-Needles) cause thin, wavy, or ``ribbon-like'' ribs, which is a different morphology.
            \item \textbf{22q11.2 deletion syndrome} is not associated with this type of severe, primary skeletal malformation.
        \end{itemize}
    \end{enumerate}
    Therefore, the investigation has now zeroed in on a specific class of disorders: those with primary, severe rib and/or vertebral malformations. My refined differential diagnosis is now heavily weighted towards these two possibilities.

    \textbf{Top-10 current differential:}
    1. Cerebrocostomandibular syndrome (CCMS), 2. Spondylocostal Dysostosis (Jarcho-Levin syndrome), 3. Atypical or severe presentation of Otopalatodigital Spectrum Disorder (FLNA), 4. COVESDEM syndrome (mandibulofacial dysostosis with vertebral/costal defects), 5. An undescribed/novel genetic syndrome involving rib development, 6. Campomelic dysplasia (if the ``gaps'' were a misinterpretation of profoundly hypoplastic ribs), 7. Diastrophic dysplasia (less likely, but a severe thoracolumbar dysplasia), 8. Stickler syndrome (now highly unlikely, only retained as a placeholder for the most common cause of PRS), 9. Larsen syndrome (severe forms can have rib abnormalities, but gaps are not typical), 10. Spondylometaphyseal dysplasia, Sedaghatian type (a lethal dysplasia with vertebral/rib defects)

    \item[Doctor 3:] Excellent pivot by Doctor2. The radiological detail of ``multiple posterior rib gaps'' combined with a ``crab-like thoracic cage'' is indeed a game-changing finding that allows us to dramatically refine our differential. As a rare disease specialist, I want to provide additional granular distinctions and ensure we're considering the full spectrum of costovertebral segmentation defects.

    \textbf{Critical Diagnostic Distinctions:}
    \begin{enumerate}
        \item \textbf{Cerebrocostomandibular Syndrome (CCMS):}
        \begin{itemize}
            \item \textbf{Classic triad:} Pierre-Robin sequence + posterior rib gaps + severe micrognathia
            \item \textbf{Rib defect:} Specific POSTERIOR gaps (ribs 5-9 typically), not vertebral anomalies
            \item \textbf{Genetics:} SNRPB mutations (autosomal recessive), though many cases remain genetically unresolved
            \item \textbf{Key feature:} The spine radiograph is typically NORMAL or near-normal
            \item \textbf{Prognosis:} Variable; respiratory compromise in infancy, but intellectual development often normal if airway managed
        \end{itemize}
        \item \textbf{Spondylocostal Dysostosis (SCD):}
        \begin{itemize}
            \item \textbf{Primary defect:} Extensive vertebral segmentation anomalies (hemivertebrae, block vertebrae, butterfly vertebrae)
            \item \textbf{Secondary effect:} ``Crab-like'' or ``fan-like'' thorax due to ribs arising from malformed spine
            \item \textbf{Multiple genetic subtypes:} DLL3, MESP2, LFNG, HES7, TBX6
            \item \textbf{Key feature:} Spine X-ray shows SEVERE vertebral malformations
            \item \textbf{Associated:} Short trunk dwarfism, but craniofacial involvement is less typical
        \end{itemize}
        \item \textbf{Important consideration - COVESDEM syndrome:}
        \begin{itemize}
            \item Stands for: \textbf{CO}stal, \textbf{VE}rtebral, \textbf{S}egmentation \textbf{D}efects with \textbf{M}esomelia
            \item Combines features of both above: vertebral anomalies + rib defects + mandibulofacial dysostosis
            \item Extremely rare but would fit this phenotype perfectly
        \end{itemize}
    \end{enumerate}

    \textbf{Practical Next Steps for Discrimination:}
    \begin{enumerate}
        \item \textbf{Lateral spine X-ray} is THE discriminating test:
        \begin{itemize}
            \item Normal/minimal vertebral anomalies $\to$ CCMS
            \item Extensive vertebral segmentation defects $\to$ SCD
            \item Both significant $\to$ COVESDEM or overlap syndrome
        \end{itemize}
        \item \textbf{Detailed rib analysis:}
        \begin{itemize}
            \item Isolated posterior gaps $\to$ CCMS
            \item Generalized rib fusion/crowding $\to$ SCD
            \item Both features $\to$ complex/severe phenotype
        \end{itemize}
        \item \textbf{Genetic testing priority:}
        \begin{itemize}
            \item SNRPB for CCMS
            \item DLL3, MESP2, LFNG, HES7, TBX6 for SCD panel
            \item Consider exome if both negative
        \end{itemize}
    \end{enumerate}

    \textbf{My final refined top-10 differential (heavily weighted by the radiological findings):}
    \begin{enumerate}
        \item \textbf{Cerebrocostomandibular syndrome (SNRPB)}
        \item \textbf{Spondylocostal dysostosis (DLL3/MESP2/LFNG/HES7/TBX6)}
        \item \textbf{COVESDEM syndrome} (costovertebral segmentation defects with mesomelia)
        \item \textbf{Spondylothoracic dysostosis} (MESP2; more uniform rib involvement)
        \item \textbf{Atypical severe ribbing syndrome} (possible novel genetic entity)
        \item \textbf{VACTERL association with prominent costovertebral defects}
        \item \textbf{Femoral-facial syndrome} (can have rib abnormalities + Robin sequence)
        \item \textbf{Otopalatodigital spectrum disorder type II} (FLNA; if ``gaps'' represent severe hypoplasia)
        \item \textbf{Cerebrocostomandibular-like syndrome, autosomal dominant} (reported rare families)
        \item \textbf{Campomelic dysplasia, severe form} (SOX9; retained as differential but now less likely)
    \end{enumerate}
    \textbf{Bottom line:} The radiological finding has moved us from a broad skeletal dysplasia differential to a highly specific costovertebral malformation spectrum, with \textbf{CCMS and SCD as the two overwhelming top candidates}. The key discriminator is the spine X-ray.

    \item[Supervisor:] The panel has reached a unanimous consensus. Based on the pathognomonic finding of posterior rib gaps and the characteristic craniofacial features, Cerebrocostomandibular syndrome is the primary diagnosis, with Spondylocostal dysostosis remaining the critical differential pending spine imaging.
    
        \textbf{Final Consensus Top-10 Differential Diagnosis:}
        \begin{enumerate}
            \item Cerebrocostomandibular syndrome
            \item Spondylocostal dysostosis
            \item Spondylothoracic dysostosis
            \item COVESDEM syndrome
            \item Melnick-Needles syndrome
            \item Campomelic dysplasia
            \item Diastrophic dysplasia
            \item Weissenbacher-Zweymüller syndrome
            \item Larsen syndrome
            \item Femoral-facial syndrome
        \end{enumerate}
    
        \textbf{TERMINATE}.
\end{description}

\end{document}